# EqSpike: Spike-driven Equilibrium Propagation for Neuromorphic Implementations

Erwann Martin[1], Maxence Ernoult[2,3], Jérémie Laydevant[2], Shuai Li[2], Damien Querlioz[3], Teodora Petrisor[1], Julie Grollier[2]

[1] - Thales Research and Technology, 91767 Palaiseau, France

[2] - Unité Mixte de Physique, CNRS, Thales, Université Paris-Saclay, 91767 Palaiseau, France

[3] - Université Paris-Saclay, CNRS, Centre de Nanosciences et de Nanotechnologies, 91120 Palaiseau, France

**Summary**

Finding spike-based learning algorithms that can be implemented within the local constraints of neuromorphic systems, while achieving high accuracy, remains a formidable challenge. Equilibrium Propagation is a promising alternative to backpropagation as it only involves local computations, but hardware-oriented studies have so far focused on rate-based networks. In this work, we develop a spiking neural network algorithm called EqSpike, compatible with neuromorphic systems, which learns by Equilibrium Propagation. Through simulations, we obtain a test recognition accuracy of 97.6% on MNIST, similar to rate-based Equilibrium Propagation, and comparing favourably to alternative learning techniques for spiking neural networks. We show that EqSpike implemented in silicon neuromorphic technology could reduce the energy consumption of inference and training respectively by three orders and two orders of magnitude compared to GPUs. Finally, we also show that during learning, EqSpike weight updates exhibit a form of Spike Timing Dependent Plasticity, highlighting a possible connection with biology.

**Introduction**

Spike-based neuromorphic systems have, in recent years, demonstrated outstanding energy efficiency on inference tasks (*1*). Implementing the training of deep neural networks in such systems remains, however, a considerable challenge, as backpropagation does not apply directly to spiking networks and requires spatially non-local computations that go against the principles of neuromorphic systems. A large number of neuromorphic systems use the unsupervised and biologically-inspired Spike Timing Dependent Plasticity (STDP) learning rule because its weight updates, based on the relative timing of pre- and post-synaptic spikes, are spatially local and can be achieved with compact circuits in several technologies (*2–11*). Unfortunately, STDP weight updates generally do not minimize a global objective function for the network, and the accuracy of STDP-trained neural networks remains below state-of-the-art algorithms based on the error backpropagation (*12*). Important research efforts therefore investigate how the error backpropagation algorithm can be mathematically modified to make it spatially local and appropriate for spiking neural networks (*13–19*). The derived learning rules are composed of three factors. The first two take into account, as usual, the behaviour of pre and post neurons, and the third allows for the introduction of an additional error factor. This third factor leads to implementations on neuromorphic chips that are less compact, and possibly less energy efficient, than two-factor learning rules such as STDP (*20*).

In this work, we propose a different approach to training spiking neural networks with high accuracy while using a local, two-factors learning rule compatible with neuromorphic implementations and scalable to complex tasks. Instead of starting from a non-local algorithm such as backpropagation and modifying it to make it local, we start from a rate-based algorithm called Equilibrium Propagation (*21*) that is intrinsically local in space, and features key advantages for neuromorphic implementations (*22*, *23*). Equilibrium Propagation theoretically applies to any physical system whose dynamics derive from an energy function. By minimizing the energy of such a system on data patterns, it can be made to relax towards states of minimal error prediction with respect to targets (*21*). The weight updates of Equilibrium Propagation match those of Back-Propagation-Through-Time (BPTT) in recurrent neural networks with static inputs (*24*), and it reaches high accuracy on image benchmarks such as CIFAR-10 (*25*). Equilibrium Propagation uses the same set of weights for the forward and backward pass, a feature that is not biologically plausible, but is interesting for neuromorphic computing as it decreases the number of synaptic devices to update, thus reducing the overall power consumption. Contrarily to backpropagation, Equilibrium Propagation uses the same computations in the forward and backward phases, which is another highly desirable feature for neuromorphic systems as it greatly simplifies the circuits. Equilibrium propagation is, however, originally a rate-based algorithm.

Here, we design a spiking, hardware-friendly version of Equilibrium Propagation, called EqSpike, compatible with current neuromorphic technologies achieving online learning (*26–32*). EqSpike is local in space and time: contrarily to backpropagation, neither error gradients nor activations need to be stored in external memories, and synapses can be directly updated through neural events. We simulate a fully connected network based on this architecture on the MNIST handwritten digits database. We obtain a test recognition accuracy of 97.6%, which compares favourably with spiking neural networks learning with backpropagation-derived methods, and on par with rate-based Equilibrium Propagation. We show that EqSpike can be implemented in silicon neuromorphic technology, and thus reduce the energy consumption of inference by up to three orders of magnitude and training by up to two orders of magnitude compared to graphics processing units (GPUs). Finally, we also show that during learning the weight updates of EqSpike exhibit a form of STDP, yielding insights to its link to biology.

**EqSpike: a hardware-friendly spiking version of Equilibrium propagation**

Equilibrium Propagation is an algorithm for training convergent recurrent neural networks. Input neurons are clamped to a static input and all the other neurons, bi-directionally connected through synapses, evolve dynamically in time to reduce the energy of the network (*21*). The algorithm functions in two phases: a free phase and a nudging phase. In the free phase, performing inference, the network is let to reach equilibrium (Fig. 1a). Once this is done, inputs are kept clamped, and output neurons are nudged towards the desired output (Fig. 1b). During this nudging phase, the prediction error at the output layer is converted into a "force" acting upon output neurons and propagating to the rest of the system through time until a second equilibrium is reached. For training, synaptic values are updated by probing the neuron states after (*21*) or during (*33*) the nudging phase through a learning rule that has been shown theoretically and numerically to match the updates of Back-Propagation Through Time, the state-of-the-art algorithm for such recurrent neural networks (*24*). It has been shown recently that Equilibrium Propagation also reaches accuracy within 1% of BPTT with convolutional architectures on the CIFAR-10 dataset (*25*).

The original version of Equilibrium Propagation uses a rate-based formulation where dynamical neurons evolve smoothly in time. For a network of leaky integrate and fire neurons described by an Hopfield-like energy function $E(u) = 1/2 \sum_i u_i^2 - 1/2 \sum_{i \neq j} \rho(u_i)\rho(u_j) - \sum_i b_i \rho(u_i)$, where $u$ are the membrane potentials of neurons and $\rho$ their activation function, the Equilibrium Propagation learning rule is: $\Delta W_{ij} \sim (\rho_i \rho_j)_n - (\rho_i \rho_j)_f$, where the product $\rho_i \rho_j$ is measured at equilibrium, at the end of the nudge phase and the free phase (*21*). This rule can be extended to the case when weights are continuously updated during the nudging phase (*33*):

$$\frac{dW_{ij}}{dt} \sim \dot{\rho}_j \rho_i + \dot{\rho}_i \rho_j, \quad \text{(Eq.1)}$$

where $W_{ij}$ is the synaptic weight connecting neurons i and j, and $\rho_i$, $\rho_j$ are the rates of the two neurons. The network dynamics thus directly compute the error derivative, encoded in the rate derivative of the post neuron $\dot{\rho}$ and multiplied by the activation function of the pre neuron. The reformulation of this rate-based learning rule to a spiking neural network is therefore the following: *each time neuron i spikes, the weight should be updated by a quantity proportional to the derivative of the rate of neuron j, $\dot{\rho}_j$ (first term in Eq. 1), and reciprocally.*

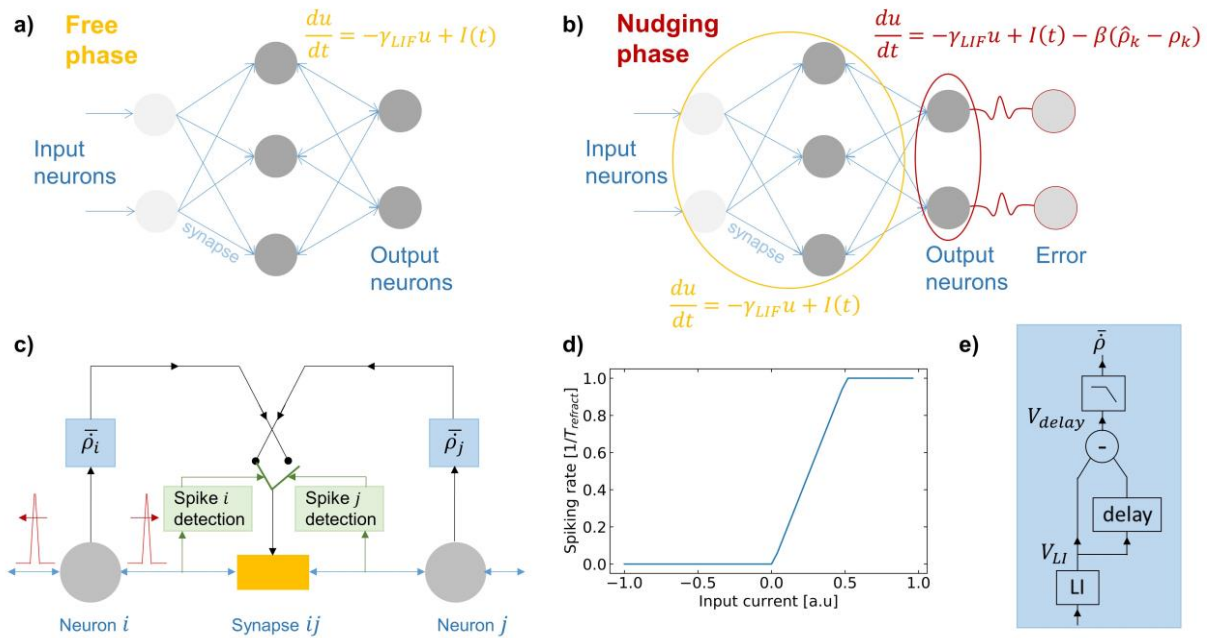

*Figure 1 - EqSpike: spike-driven Equilibrium Propagation a) Schematic of the free phase in Equilibrium Propagation. b) Schematic of the nudging phase in Equilibrium Propagation. c) Illustration of the weight update implementation in EqSpike. d) Spiking rate of the neuron as a function of the amplitude of the input signal. e) Schematic of the rate acceleration computation.*

We propose here a simple strategy, compatible with current electronic hardware, to implement this learning rule. It is illustrated in Fig. 1c in the form of a circuit, including spike detection elements at the output of each neuron, as well as dedicated blocks that extract the rate derivative from the spike trains of each neuron in real-time, in order to update synapses accordingly.

We use leaky-integrate-and-fire (LIF) spiking neurons which output spike frequency as a function of input current approximates the hard sigmoid prescribed in the original formulation of Equilibrium Propagation (*21*). Their maximum frequency is $f_{max} = 1/T_{refract}$, where $T_{refract}$ is the refractory time of the LIF neuron (see Supplementary Information for details and all parameter values).

The novelty compared to standard SNNs is the scheme that we propose for extracting the rate acceleration $\dot{\rho}$ for each neuron, illustrated in Fig. 1e. A leaky-integrator with a leak factor $\gamma_{LI}$ (without reset nor spikes), takes as input the spike train emitted by the neuron to which it is connected and outputs a slowly varying signal proportional to the rate of the neuron spike train: $V_{LI} \sim \rho/\gamma_{LI}$ (35). To take the derivative, we delay this signal by a duration $\tau$, and subtract the actual value with the delayed value:

$V_{delay} = V_{LI}(t) - V_{LI}(t - \tau) \cong \tau \frac{\partial V_{LI}}{\partial t} \propto \tau/\gamma_{LI} \dot{\rho}$. We then apply a low-pass filter for smoothing the variations. The filter is simulated using an average over $N_{filt}$ simulation steps: $\overline{x(t)} = \frac{1}{N_{filt}} \sum_{i=0}^{N_{filt}-1} x_i(t - i\, dt)$, where $dt$ is the simulation time step.

The output of the filter, that approximates $\frac{\tau}{\gamma_{LI}} \bar{\rho}$, is then multiplied by the coefficient $\eta_r$. The corresponding weight updates are $\Delta w_{ij} = \eta_r \frac{\tau}{\gamma_{LI}} \bar{\rho}_i$, which corresponds to an effective learning rate $l_r = \eta_r \frac{\tau}{\gamma_{LI}} = 1.5\, 10^{-3}$.

This approach is hardware-compatible as LIF neurons, leaky integrators, delays and low pass filters are circuit elements that can be efficiently implemented in CMOS technology (*36*), and bidirectional synapses could be implemented with CMOS compatible emergent nano-devices such as memristors (*31*, *37*, *38*). The corresponding pseudo-code is given in Algorithm 1 (see Supplementary Information for details).

---

**Algorithm 1:** EqSpike learning procedure for one image

**Inputs**: input image, Model ($n_{inputs}, n_{hidden}, n_{out}$), Loss function, length Free phase $T_{free}$, length Nudging phase $T_{nudge}$, Parameters $\gamma_{LIF}, \gamma_{LI}, u_{th}, \beta, \eta_r, \tau, N_{filt}, W_{ij}$

for $t < T_{free}$:                                                       ■ free phase
  for each neuron $j$:
    Update membrane potential $u_j(\gamma_{LIF}, I_j)$
    if $u_j > u_{th}$:
      Emit a spike ($t_j$)
    Update $\rho_j(t_j, \gamma_{LI})$

for $t \in [T_{free}, T_{free} + T_{nudge}]$:                              ■ nudging phase
  for each output neuron $o$:
      Compute error gradient $\nabla e_o$
      Nudge neuron: $u_o \leftarrow u_o - \beta \cdot \nabla e_o$
  for each neuron k:
    Update $u_k(\gamma_{LIF}, I_k)$
    if $u_k > u_{th}$:
      Emit a spike ($t_k$)
    Update $\rho_k(t_k, \gamma_{LI})$
    Compute smoothed: $\bar{\rho}_k\left( (\rho_k(t_k), \rho_k(t_k - \tau)), \ldots, N_{filt} \right)$

  for each synapse $w_{ij}$:                                      ■ Update synapses
    if neuron j emits a spike:
      $w_{ij} \leftarrow w_{ij} + \eta_r \cdot \frac{\tau}{\gamma_{LI}} \bar{\rho}_i$
    if neuron i emits a spike:
      $w_{ij} \leftarrow w_{ij} + \eta_r \cdot \frac{\tau}{\gamma_{LI}} \bar{\rho}_j$

**Return**: Trained weights for input image: $W_{ij}$ and go to next image/next epoch.

**Full network simulations: recognition rate on handwritten digits database**

We now evaluate the performance of EqSpike on the MNIST handwritten digits classification task, using a fully connected network with one hidden layer (see Supplementary Information for details). The obtained train (orange) and test (blue) accuracies are shown Fig. 2 as a function of the number of training epochs, with the deviation over six runs in shadow color. Table 1 compares the results to BPTT and the version of Equilibrium Propagation closest to our implementation, called Continual Eq-Prop (*33*), trained with a batch size of one.

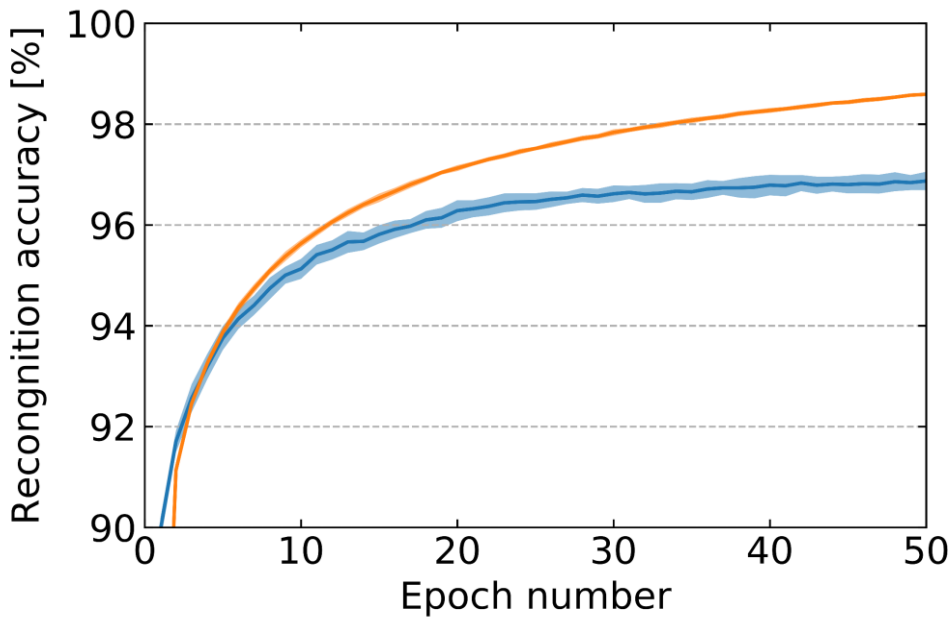

*Figure 2 - Recognition accuracy during training (orange) and test (blue) as a function of the number of epochs for MNIST, averaged over six runs.*

|  | **BPTT** 784-100-10 | **Continual Eq-Prop** 784-100-10 | **EqSpike 100** 784-100-10 | **EqSpike 300** 784-300-10 |
|---|---|---|---|---|
| MNIST | Test: **97.11%** ± 0.23% Train: 99.06% ± 0.15% (average on 6 runs) | Test: **96.97%** ± 0.12% Train: 99.8% ± 0.04% (average on 6 runs) | Test: **96.87%** ± 0.18% Train: 98.59% ± 0.03% (average on 6 runs) | Test: **97.59%** ± 0.1% Train: 98.91% ± 0.03% (average on 6 runs) |

*Table 1 - Comparison between BPTT, C-EP and EqSpike,* with the same initialization procedure. Batch size = 1.

The test accuracy of EqSpike matches closely the accuracy of stochastic gradient descent through BPTT on the same network architecture, given the error margin. With a hidden layer of 300 neurons, EqSpikes reaches a test accuracy of 97.59%. Fully-connected spiking neural networks trained on MNIST without conversion from a non-spiking neural network typically achieve recognition rates in the 96-98% range (*13*, *39–42*). EqSpike, with its local, two-factors online learning rule therefore reaches accuracies on MNIST comparable to those of the latest models investigated for training spiking neural networks on hardware platforms. We have chosen the MNIST dataset as a benchmark because it is a standard dataset for the neuromorphic community interested in training spiking neural networks online, due to the long simulation times. As EqSpike achieves results on MNIST equivalent to the baseline given by Equilibrium Propagation it has the potential, like Equilibrium Propagation, to adapt to convolutional architectures and perform with good accuracy on more complex image benchmarks (*25*), with the additional advantage of being compatible with current neuromorphic technologies.

**Inference speed and energy**

As EqSpike is derived from a rate-based algorithm, it is interesting for neuromorphic applications to quantify the number of spikes needed to achieve inference, and the time needed to reach high accuracy. Operation with fewer spikes is more desirable, as it reduces both execution time and energy consumption.

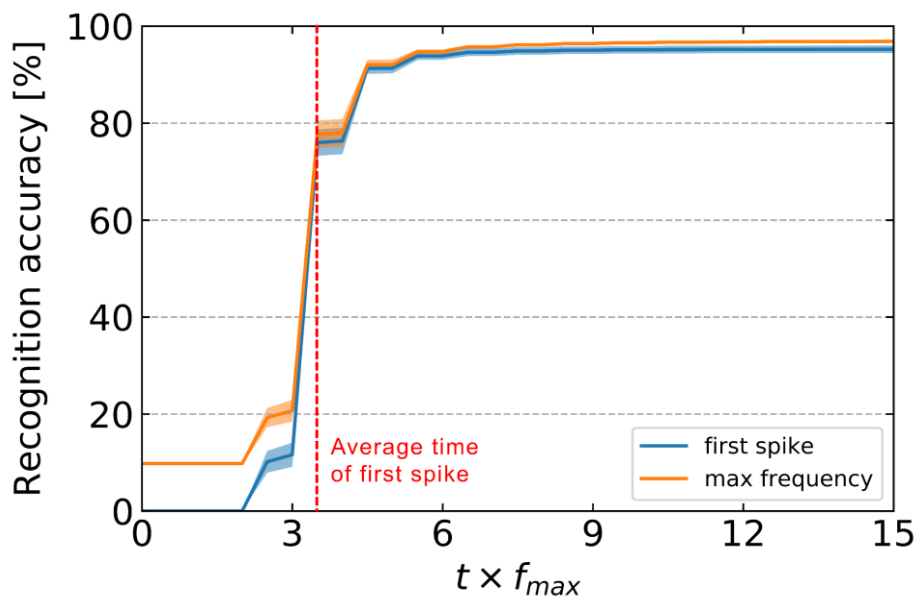

*Figure 3 - Inference time: recognition accuracy on MNIST on the test dataset as a function of time multiplied by the maximum neuron frequency $f_{max}$. Orange line: recognition accuracy computed from the output neuron showing the highest rate. Blue line: recognition accuracy computed from the output neuron spiking first. Red, vertical dotted line: average time of first output spike.*

Fig. 3 shows the inference results as a function of the execution time multiplied by the maximum frequency of neurons ($t \times f_{max}$). The orange line is the mean accuracy result over the whole test dataset, obtained by computing the spike rate of output neurons (as done for Fig. 2 and Table 1), through averaging in a time window $T_{average}$ of 100 simulation time steps ($T_{average} = 50/f_{max}$) and considering that the neuron with highest frequency encodes the output. This method computes the rate accurately at the expense of having to wait the time $T_{average}$ and letting output neurons spike multiple times. Spiking neural networks also offer the possibility to accelerate the computation and reduce the energy consumption by determining the output class from the first output neuron to spike. The blue line in Fig.3 is the accuracy as a function of time, averaged over all images in the test dataset, obtained by considering that the first output neuron to spike encodes the output.

The red, vertical dotted line in Fig. 3 indicates the average time of the first spike at the output over all presented images, corresponding to t ≅ $3.5/f_{max}$. At t ≅ $10/f_{max}$, the accuracy of single-spike inference (blue curve) reaches 95.11% ± 0.78%, within 1.4% of the precise rate computation (orange curve). This result shows that even though the algorithm is originally rate-based, a single spike at the output suffices in most cases to determine the correct class with good precision, a feature which is highly attractive for energy-efficient inference on neuromorphic chips. It means that inference can be achieved in 100 μs for electronic neurons with a firing rate of 100 kHz, available in neuromorphic chips working in accelerated time compared to biology (*26*), and 1 μs for electronic neurons with a firing rate of 10 MHz that can be produced, for example, with emerging nanotechnologies (*43*). The corresponding throughputs are respectively 10k and 1M images/s, on par with current spiking neural network implementations (*32*, *44*). As the network operations are fully parallel, these orders of magnitudes will be conserved for wider networks. Simulations of rate-based Equilibrium Propagation on deeper networks indicate that the convergence time increases by a factor of about eight for a network with four hidden layers compared to a network of one hidden layer as here (*25*). It should be noted that in the current implementation we present static inputs to the network, which means that input neurons need to integrate these signals before they emit the first spikes that will then propagate to the next layers. The speed of inference could be increased in the future by presenting inputs directly encoded in spikes, for example sourced from neuromorphic vision sensors (*44*).

An estimation of the energy consumption of a spiking neural network on a neuromorphic silicon chip can be performed by counting the number of synaptic operations involved. Synaptic operations (SynOps) are defined as the total number of spikes transiting through synapses of the network. Frenkel et al show that a SynOp on a neuromorphic chip requires as little as 10 pJ (*30*). The total number of synaptic operations needed for inference depends on the targeted recognition precision and, therefore, on the duration of inference (Fig. 3). For EqSpike, the recognition rate saturates at t ≅ $10/f_{max}$. The corresponding measured number of SynOps is about 150,000 in average. This is much less than expected if all neurons spiked. This is also a bit less but comparable to the number of SynOps needed at inference for Event-Driven Random BackPropagation (*13*). Considering 10 pJ/SynOps, each EqSpike inference could potentially consume 1.5 μJ. This means that testing the 10,000 images of the whole MNIST dataset could be achieved with a neuromorphic chip while consuming only 15 mJ, in other words, three orders of magnitude less than with a GPU (*45*).

In our current EqSpike implementation, the input layer is the one leading to most spikes and SynOps: with only 16% of illuminated pixels in average in MNIST, the input layer emits 87.5% of all spikes and 98.6% of SynOps occur between the input layer and the hidden layer. In this work, we did not focus on reducing the number of spikes with encoding, but a better encoding of the input may reduce considerably the energy consumption. Kheradpisheh et al have shown that with a temporal encoding, the total number of spikes in the network before the first output spike can be reduced to 200 with a hidden layer four times larger than our network (*46*). In our case, 678 spikes in total are emitted in average before the first output spike. An adaptation of EqSpike to temporal encoding is not straightforward, but this number could potentially be decreased in the future by reducing the encoding frequency of the input.

**Training speed and energy**

Training with EqSpike requires performing the free phase and then the nudging phase, during which synaptic weights are updated. A way to speed-up the training, and reduce the total number of SynOps, is to perform the nudging phase only on poorly classified examples, and skip the updates when the accuracy is satisfactory. We apply this strategy inspired from (*32*) using the criteria that the nudging phase is performed only when the difference between the target rate and the actual rate ($\widehat{\rho_k} - \rho_k$) at the ouput is above 1%. Fig. 4a shows the number of presented examples per epoch as a function of epoch number. In the last 20 epochs only approximately 15% of the training dataset still require a nudging phase.

For MNIST, we thus perform the free phase on all the dataset ($3\times10^6$ images for the 50 epochs), and the nudging phase on 489,000 images. Given the durations of each phase, we can estimate the training time to $T_{training} \cong 2.74\times10^8/f_{max}$. For electronic neurons with a firing rate of 100 kHz (*26*), this leads to $T_{training} \cong 45$ min, and for electronic neurons with a firing rate of 10 MHz (*43*) to $T_{training} \cong 30$s. As our networks feature a fully parallel nature, these training times would be the same for much wider networks, and increased by a factor of about eight only with four hidden layers (*25*).

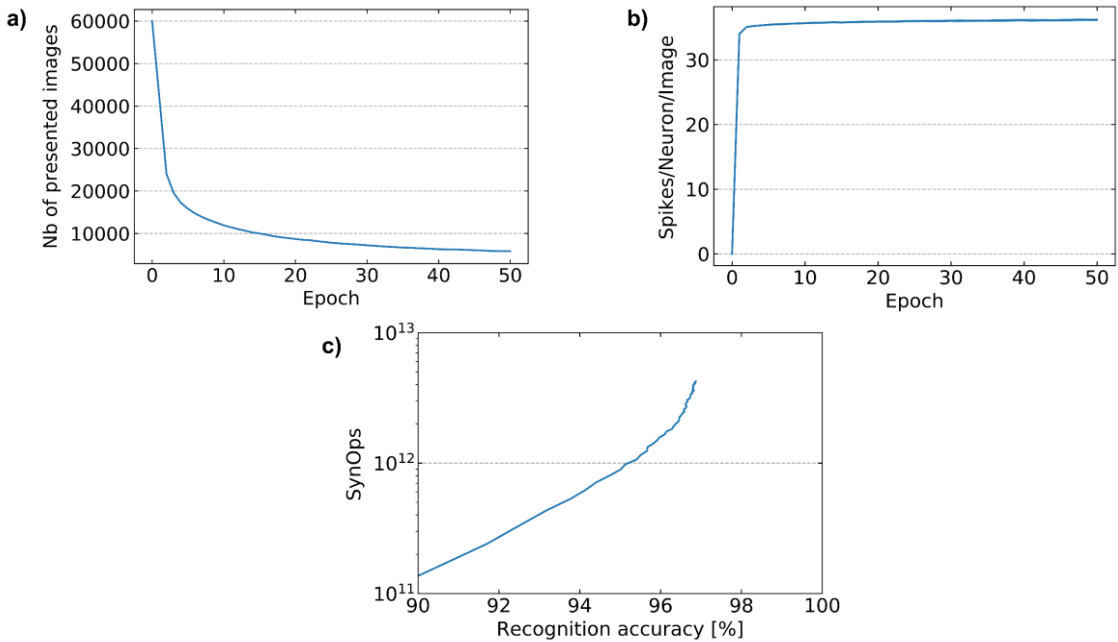

**Figure 4 - Training performance** a) Number of presented images in the nudging phase per epoch versus epoch number. b) Number of spikes/neuron/image occurring during the two phases (nudge+free), as a function of the epoch. c) SynOps: number of spikes during both phases (nudge+free) as a function of the recognition accuracy.

As EqSpike is derived from a rate-based approach, it is interesting to compare the actual spiking rates of neurons in the network during training to their maximum frequency $f_{max}$. For neuromorphic applications, low overall rates are indeed desirable. Fig 4b shows the average number of spikes emitted by each neuron for an image presentation in the training dataset, as a function of the epoch. We found that for the training conditions of Fig. 2, there are in average 36 spikes/neuron/image. This means that neurons in the network spike in average with a frequency of the order of 20% of $f_{max}$, well below $f_{max}$, which is promising for neuromorphic implementations. Again this number could be reduced in the future by optimizing the encoding of input at the first layer.

Fig. 4c shows the numbers of synaptic operations needed for training as a function of recognition rate. The total number of synaptic operations after 50 epochs is of $4.23\times10^{12}$, which is of the same order of magnitude as Event-Driven Random Back-Propagation (*13*) for similar accuracy, and below training

MNIST with BP based on (14). With 10pJ per SynOps (30) the training phase of EqSpike on a neuromorphic chip could consume as little as 42 J, again, two orders of magnitude less than with a GPU (45).

**Spike Timing Dependent Plasticity**

We have shown that EqSpike transforms Equilibrium Propagation into an efficient algorithm for neuromorphic chips. We now highlight that it also brings Equilibrium Propagation closer to biological plausibility. Bengio et al have pointed out a connection between the Equilibrium Propagation learning rule of Eq. 1 and STDP (47). The STDP learning rule, illustrated in Fig. 5a, reinforces causality between the spikes of pre- and post-synaptic neurons in networks with unidirectional synapses. If the post-synaptic neuron spikes after the pre-, causality is observed, and the weight is increased. In the opposite scenario, the weight is decreased.

Let us consider a situation where the pre-synaptic neuron i spikes and the post-synaptic neuron j accelerates, as illustrated in Fig. 5b. According to the Eq-Prop learning rule, in a network with unidirectional synapses, $\frac{dW_{ij}}{dt} \propto \dot{\rho}_j \rho_i = \dot{\rho}_{post} \rho_{pre}$, a positive weight update should be applied. Due to the acceleration of the post-neuron, there are less post-neuron spikes before the pre-neuron spike than after. Therefore, $t_{post}$ - $t_{pre}$ is positive in average, yielding a positive weight update through STDP.

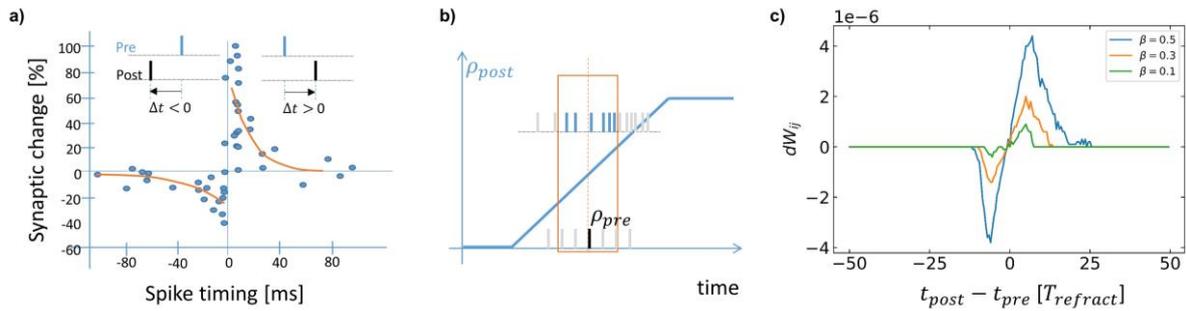

*Figure 5 - STDP.* a) Illustration of the STDP learning rule; reproduction with data from (2). b) Illustration of the link between Eq-Prop and STDP learning rules; illustration reproduced from (47). c) STDP-like curve during EqSpike learning.

We have investigated if STDP-like weight updates did emerge during learning in our simulations. For this purpose, we monitored weight variations in synapses that connect input neurons and the hidden layer neurons. These synapses are unidirectional as input neurons are clamped to the input (see Fig. 1): their frequency does not vary. We used 100 images during the first epoch for the MNIST dataset. The curve in Fig. 5c shows the average weight updates in the first layer as a function of the time difference between post-synaptic neuron spikes and the average time of pre-synaptic neuron spikes in a window of 200 time steps before the post-synaptic neuron spike. The obtained curve, centred on zero, indeed exhibits an STDP-like shape. It has been obtained by filtering out very low frequencies below 0.05. Quiet neurons in the free phase indeed induce large weight updates at the beginning of the nudging phase, due to the sudden acceleration from zero to non-zero frequency, inducing an additional noise in the curve. It should be noted that biological STDP curves being frequency-dependent, they are also often obtained by focusing on a given range of frequencies (48). Fig. 5c shows that the STDP amplitude and time-window vary with the strength of nudging $\beta$. In Equilibrium Propagation, weight changes are driven by neuron accelerations. As nudging applies a force that accelerates neurons, for larger $\beta$, larger weight changes are obtained, leading to a larger amplitude of the STDP-like curve. As higher accelerations are reached during the nudging phase, noticeable weight modifications occur for larger differences between $t_{post}$ and $t_{pre}$, corresponding to a wider time window for the STDP-like curve. These results confirm the possible connection between STDP and Equilibrium Propagation pointed out in (*47*), despite the fact that individual spike timings are lost through averaging. They show that STDP-like behaviour can be obtained during learning in unidirectional synapses without a direct implementation of the original, causality-based rule. They also ask the question whether multilayer networks could be trained with the local STDP learning rule by using the nudging procedure of Equilibrium Propagation, possibly in the extended version called "Vector-field Equilibrium Propagation" in which synapses are unidirectional (*49*).

**Discussion**

Other versions of Equilibrium Propagation have been specifically designed to train spiking neural networks. O'Connor et al have designed a spiking network which, by construction, stochastically approximates the dynamics of its rate-based counter-part until reaching the same steady state with a minimal spikes communication budget between neurons (*50*). Their technique, successfully tested against MNIST, comes at the cost of combining predictive coding, sigma-delta modulation and adaptive step sizes at the neuron level. In comparison, our technique simply requires LIF neurons, a spike-count and a low-pass filter to implement the learning rule. Mesnard et al proposed a version of Equilibrium Propagation to train spiking networks that also makes use of low-pass filters to estimate firing rates. However, they only demonstrate their approach on a non-linear toy problem, and the implemented learning rule is not local in time (*51*). In contrast, we demonstrate the effectiveness of our fully event-based implementation on MNIST.

More generally, several spike-based approaches to backpropagation have been proposed. One method consists in smoothing the spikes as a function of time so that gradients can be back-propagated through time (*18*, *52*). Another technique consists in gating the spikes propagating the error signals by surrogate derivatives, as done in SpikeGrad (*53*). Finally, Event-driven Random Backpropagation uses firing rates in the backward pass, as we do in this paper, and achieves similar performance on MNIST (*13*). In comparison with these approaches however, our implementation of Equilibrium Propagation does not require to know the activation function equation in order to compute its derivatives. One other interesting approach is S4NN, which employs latency coding, where each neuron can spike at most once, and the output is encoded as the first neuron to spike (*46*). In contrast with rate-based coding, latency coding has the potential to save important energy in neuromorphic implementations. However, there is no clear indication of whether S4NN could scale to harder visual tasks, while rate-based Equilibrium Propagation was shown to train deep ConvNets on CIFAR-10 (*25*). One intrinsic limitation of EqSpike, however, may stem from the necessity for the system to reach a steady state before the gradient computation phase. This prevents, for now, the classification of time-varying inputs. Payeur et al recently proposed a spike-based approach where top-down error signals are encoded as spike bursts and are multiplexed with bottom-up feedforward signals so that the backward pass and the forward pass can occur simultaneously with minimal disruption (*19*). While they show that the rate-based counterpart of their algorithm works on CIFAR-10 and ImageNet, it comes at the cost of employing dendritic network topologies and specialized synapses. Indeed, as noted in the introduction, most backpropagation-derived local learning rules involve a third factor for supervision (*20*). In this regard, we believe that our EqSpike implementation of Equilibrium Propagation, with only two factors, achieves an optimal trade-off between circuitry complexity and performance.

Finally, Zoppo et al (*22*) and Kendall et al (*23*) have proposed using Equilibrium Propagation for training neuromorphic hardware. However, their implementations remain either rate-based or current-based, and are therefore not directly compatible with spiking neuromorphic chips. EqSpike, on the other hand, could be trained directly on reconfigurable neuromorphic systems with online learning such as SpiNNaker (*27*) and Loihi (*29*).

**Limitations of the study**

A limitation of EqSpike is that, with a wrong choice of $\gamma_{LI}$ or $\tau$ combined with low nudging times, fast or brief rate changes cannot be detected, and the corresponding weight modification cannot be applied. This does not seem to be a problem with our experimentations as demonstrated above but could lead to a lower accuracy on more complex problems.

The circuits to compute the rate derivative could be further miniaturized by extracting $\dot{\rho}$ directly from the membrane potential if neuron models with smoothly varying membrane potential are used (*34*).

EqSpike could also be sped up by building on dedicated hardware in analog or digital CMOS (*10*, *26*, *28*, *30*, *32*). Emerging nanotechnologies such as memristive synapses and nanoscale spiking oscillators are compelling candidates to scale up neuromorphic hardware due to their small size, their speed and their low energy consumption (*37*, *54–57*). These technologies are typically prone to imperfections such as the device-to-device variability, cycle-to-cycle variability or the non-linearity in the conductance-to-voltage response, which are known to considerably jeopardize learning in memristive neural networks (*31*, *58*). Our paper implicitly assumes that the underlying memory technology at use would be linear, deterministic and identical across different synapses. The learning rate $l_r$ in EqSpike is indeed about 10 times smaller than typically used in standard gradient descent. This is due to the fact that the batch size is one, and also that a low $l_r$ is needed to avoid too large modifications accumulated over all the nudging phase. For low precision device as synapse, a first possible solution is to update the synaptic devices less frequently (not at each spike), but with a higher value. Another strategy is to adapt to EqSpike the training schemes used for binary neural networks (*59–61*). Further study should be done to propose a fully end-to-end circuit to implement EqSpike and investigate its resilience to the memristive device imperfections mentioned.

**Conclusion**

In this work we present a new algorithm for spiking neural networks, EqSpike, compatible with neuromorphic systems, and achieving good performance on MNIST. We show that EqSpike implements the learning rule of Equilibrium Propagation locally and autonomously. The gradients are computed by the dynamics of the system and the weights are modified by a spike and the addition of only one block to the neuron. This can lead to spiking neuromorphic systems that do not need an external circuit to compute the error gradients given by backpropagation and learn autonomously, simply by presenting inputs and nudging the outputs according to errors. Our method obtained results on MNIST close to backpropagation through time and Equilibrium Propagation, two state-of-the-art algorithms. Moreover, because EqSpike is based on Equilibrium Propagation, the performance on more complex task, like CIFAR-10, could be similar. The number of synaptic operations to obtain these results in MNIST show we can obtain theoretically two orders of magnitude less energy consumption than a GPU for training and the same magnitude of time with high frequency neurons. The inference, after training the network, can be accelerated by waiting for the first output spike rather than compute the highest rate, with only a small loss of accuracy.

Finally, we show that the weight updates of EqSpike share similarity with STDP during learning, raising the question of a possible biological plausibility of the algorithm. In average, the modification of weight is proportional to the spike timing. This could permit to implement synapses in neuromorphic hardware by emergent nano-devices with a STDP-like behavior, to obtain a lower power consumption and higher surface density.

**Author contributions**

JG, TP and DQ have devised and supervised the study, EM has performed the spiking simulations with the help of ME, JL and SL. ME and JL have performed the benchmark rate-based simulations. All authors have contributed to writing the article.

**Acknowledgements**

This publication has received funding from the European Union's Horizon 2020 research innovation programme under grant agreement 732642 (ULPEC project) and 876925 (ANDANTE project)". E. Martin received funding from the ANRT grant 2018/0884.

**Declaration of interests**

The authors declare no competing interests.

**Resource availability**

*Lead contact*

Further information and requests for resources and reagents should be directed to and will be fulfilled by the lead contact, Julie Grollier (julie.grollier@cnrs-thales.fr).

*Materials availability*

N.A.

*Data and code availability*

The data generated by this study is available on reasonable request.